\documentclass[runningheads]{llncs}

\usepackage{eccv}

\usepackage{eccvabbrv}

\usepackage{graphicx}
\usepackage{booktabs}
\usepackage{multicol}
\usepackage{multirow}
\usepackage{adjustbox}
\usepackage{colortbl}
\usepackage{diagbox}
\usepackage[accsupp]{axessibility}  

\definecolor{Gray}{gray}{0.9}
\newcommand{\bihan}[1]{\textcolor{black}{{#1}}}

\usepackage{hyperref}

\usepackage{orcidlink}
\DeclareMathOperator*{\argmin}{argmin}

\begin{document}

\title{Temporal As a Plugin: Unsupervised Video Denoising with Pre-Trained Image Denoisers} 

\titlerunning{Unsupervised Video Denoising with Pre-Trained Image Denoisers}


\author{Zixuan Fu\orcidlink{0009-0005-5613-3624} \and Lanqing Guo\orcidlink{0000-0002-9452-4723} \and Chong Wang\orcidlink{0009-0004-6425-7232} \and Yufei Wang\orcidlink{0000-0002-6326-7357} \and Zhihao Li\orcidlink{0000-0002-2066-8775} \and\\ Bihan Wen\orcidlink{0000-0002-6874-6453
}\thanks{Corresponding author.}}

\authorrunning{Fu et al.}

\institute{Nanyang Technological University, Singapore\\
\email{\{zixuan006, lanqing001, wang1711, yufei001, zhihao.li, bihan.wen\}@ntu.edu.sg}}

\maketitle
\begin{abstract}
Recent advancements in deep learning have shown impressive results in image and video denoising, leveraging extensive pairs of noisy and noise-free data for supervision.
However, the challenge of acquiring paired videos for dynamic scenes hampers the practical deployment of deep video denoising techniques.
In contrast, this obstacle is less pronounced in image denoising, where paired data is more readily available.
Thus, a well-trained image denoiser could serve as a reliable spatial prior for video denoising.  
In this paper, we propose a novel unsupervised video denoising framework, named ``\textbf{T}emporal \textbf{A}s a \textbf{P}lugin'' (TAP), which integrates tunable temporal modules into a pre-trained image denoiser.
By incorporating temporal modules, our method can harness temporal information across noisy frames, complementing its power of spatial denoising.
Furthermore, we introduce a progressive fine-tuning strategy that refines each temporal module using the generated \textit{pseudo clean} video frames, progressively enhancing the network's denoising performance.
Compared to other unsupervised video denoising methods, our framework demonstrates superior performance on both sRGB and raw video denoising datasets.
Code is available at \href{https://github.com/zfu006/TAP}{https://github.com/zfu006/TAP}.

  
\end{abstract}

\section{Introduction}
\label{sec:intro}
Video denoising is one of the fundamental tasks in low-level vision which aims to restore noise-free videos from their degraded counterparts. 
\bihan{Recent advancements in deep learning-based methods have achieved great progress \bihan{in this task}, by leveraging a large number of paired noisy and clean videos for training networks such as convolutional neural networks~\cite{yue2020supervised, maggioni2021efficient, tassano2020fastdvdnet, vaksman2021patch} and transformers~\cite{liang2022vrt, song2022tempformer}.}
\bihan{However, the practical application of these methods is often hindered by the challenge of collecting paired videos, especially in dynamic scenes~\cite{yue2020supervised}. }
\bihan{Despite the existence of real video datasets~\cite{yue2020supervised, chen2019seeing}, they are often limited in size or oversimplified settings, resulting in poor generalization of the trained networks.}

\begin{figure}[!t]{}
    \centering
    \includegraphics[width=.9\linewidth]{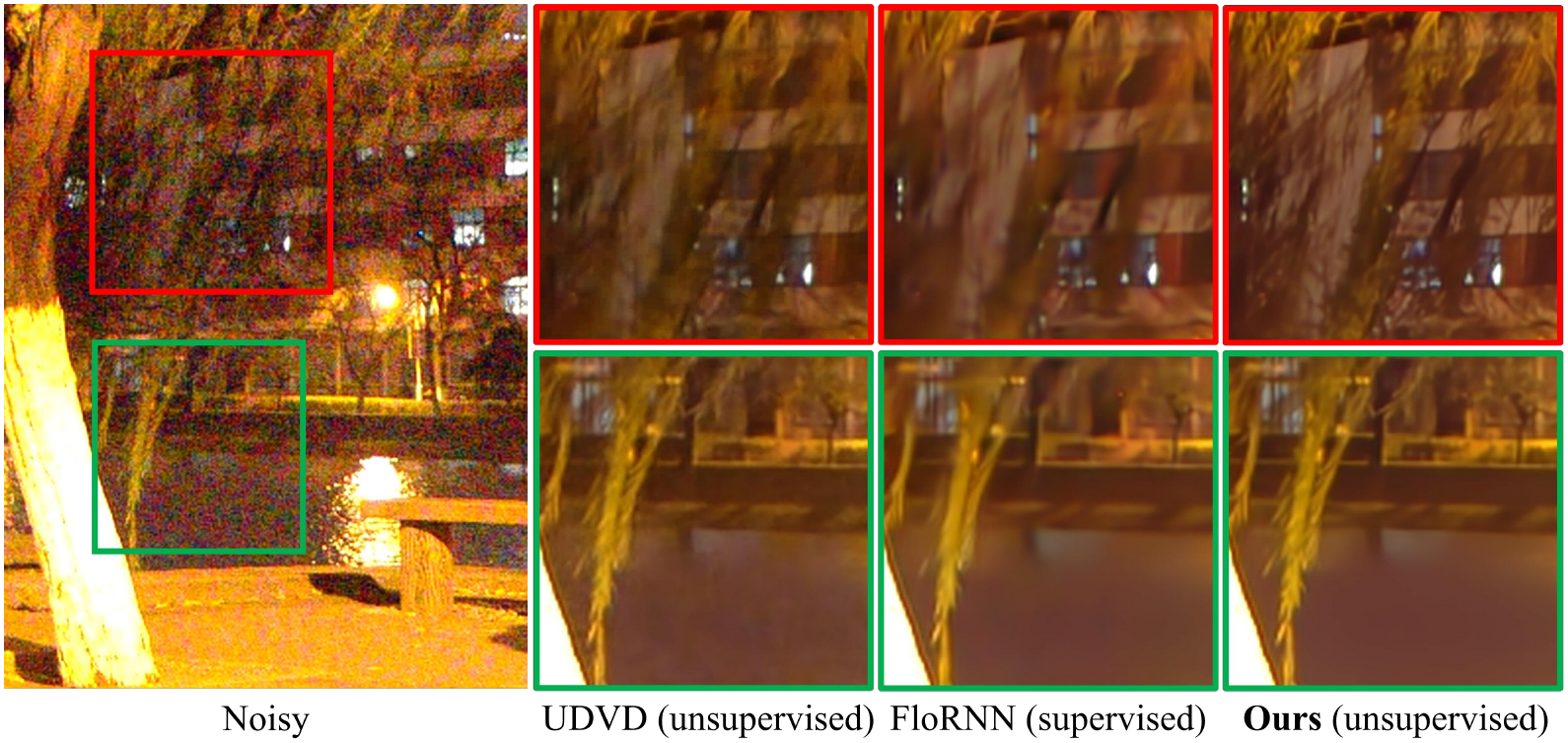}
    \caption{Examples of denoising results from the CRVD outdoor dataset~\cite{yue2020supervised}. UDVD~\cite{sheth2021unsupervised} is an unsupervised denoiser trained using noisy videos only, which suffers from remaining noise and artifacts; FloRNN~\cite{li2022unidirectional} is a supervised video denoiser, generating over-smoothing results; Our method effectively eliminates both noise and artifacts while preserving image details.
    }
    \label{fig_first_compare}
\end{figure}

\bihan{To address this issue, several unsupervised video denoising methods~\cite{dewil2021self, sheth2021unsupervised, wang2023recurrent, zheng2023unsupervised, lee2021restore} have been proposed, providing an alternative approach by bypassing the need for extensive paired videos.}
Building on previous unsupervised image denoising frameworks such as Noise2Noise~\cite{lehtinen2018noise2noise} and blind spot networks~\cite{krull2019noise2void, laine2019high, wu2020unpaired}, these methods achieve promising performance by solely utilizing noisy videos for training. 
\bihan{However, lacking corresponding guidance for supervision, their denoised results often suffer from artifacts and remaining noise~\cite{krull2019noise2void} (See examples in Fig.~\ref{fig_first_compare})}.

In this paper, \bihan{departing from previous solutions, we propose to leverage a pre-trained image denoiser as the initial spatial prior for unsupervised video denoising.} 
\bihan{Unlike supervised video denoisers whose effectiveness hinges on the availability of paired videos~\cite{dewil2021self, sheth2021unsupervised, wang2023recurrent, zheng2023unsupervised}, image denoisers can be sufficiently trained using extensive image pairs, as obtaining paired static images in real-world scenarios is more feasible~\cite{abdelhamed2018high, chen2018learning, plotz2017benchmarking, zhang2022idr, wei2020physics, nam2016holistic, xu2018real}.}

\bihan{To complement the pre-trained spatial denoising prior, we propose a novel unsupervised video denoising framework, named ``\textbf{T}emporal \textbf{A}s a \textbf{P}lugin'' (TAP).}
\bihan{To be specific, trainable temporal modules are inserted within the skip connections of an encoder-decoder based image denoiser.}
Each temporal module enables the network to incorporate temporal information on top of its learned spatial information.
Inspired by~\cite{wang2019edvr, yue2020supervised}, the arrangement of temporal modules follows the pyramid structure, allowing us to utilize the multi-scale architecture of the pre-trained network for efficient feature alignment.
Based on the built video denoiser, we propose an unsupervised fine-tuning strategy that progressively trains temporal modules from the network's bottom (deep) level to the top (shallow) level solely on noisy videos. 
The process begins with constructing pseudo noisy-clean video pairs, which involves initially denoising the noisy frames using the pre-trained image denoiser and subsequently simulating noise based on a known noise model.
We then fine-tune the temporal modules by the generated pseudo pairs while keeping the spatial layers frozen, starting from learning the temporal alignment of deep features at the bottom level to refining the structural details at the top level. This progressive strategy effectively prevents the network from overfitting to the artifacts and over-smoothness from initial pseudo labels.
We demonstrate the effectiveness of the proposed method on sRGB and raw video benchmarks. Extensive results show our proposed framework achieves state-of-the-art video denoising performance, compared to other unsupervised methods.

In summary, our main contributions are outlined below:
\begin{itemize}
    \item 
    We propose TAP, an unsupervised video denoising framework that leverages a pre-trained image denoiser for video denoising without utilizing paired videos for supervision.
    \item
    We introduce an unsupervised fine-tuning strategy that harnesses temporal information to complement the network's spatial prior, progressively enhancing its denoising performance.
    \item
    Our approach surpasses current unsupervised denoisers on both raw and sRGB video denoising benchmarks, demonstrating its superior effectiveness.
\end{itemize}

\section{Related Work}
\subsection{Deep Image Denoising}
Recently, numerous image denoisers are proposed to effectively reconstruct clean images from their noisy observations~\cite{zhang2017beyond, liu2018non, chang2020spatial, cheng2021nbnet, liang2021swinir, chen2022simple, zamir2022restormer, wang2022uformer}. 
DnCNN~\cite{zhang2017beyond}, a widely used denoiser, employs stacked plain blocks of convolution, Batch Normalization~\cite{nair2010rectified} and Rectified Linear Unit~\cite{ioffe2015batch} (ReLU), as well as global residual connection~\cite{he2016deep} for efficient denoising. 
Encoder-decoder based architectures incorporated with attention mechanisms and dynamic kernels have also been proposed to exploit local and non-local information within images~\cite{liu2018non, chang2020spatial, cheng2021nbnet}. 
Concurrently, some networks have adopted transformer-based architecture \cite{liang2021swinir, wang2022uformer, zamir2022restormer, wang2024progressive}, which achieves promising performance not only on image denoising but also on general image restoration tasks. 
Meanwhile, efforts like NAFNet \cite{chen2022simple} focus on finding simple but effective basic blocks for image restoration.  

\subsection{Deep Video Denoising}
\subsubsection{Supervised Video Denoising.} 
In contrast to image denoising, which only requires exploiting spatial information, video denoising also benefits from effective temporal alignment. 
Several supervised video denoisers are designed to effectively learn temporal information from multiple frames \cite{wang2019edvr, mildenhall2018burst, tassano2020fastdvdnet, yue2020supervised, vaksman2021patch, maggioni2021efficient, liang2022vrt, xue2019video, li2022unidirectional, liang2022recurrent, chan2021basicvsr, chan2022basicvsr++, dudhane2023burstormer}.
FastDVDNet~\cite{tassano2020fastdvdnet} learns temporal information by directly taking concatenated sequential images as inputs.
EDVR~\cite{wang2019edvr} and RViDeNet~\cite{yue2020supervised} achieve temporal alignment in the feature level through a pyramid alignment module~\cite{wang2019edvr}. 
This module employs deformable convolution for aligning adjacent features to the reference feature hierarchically. 
Besides deformable convolution, other methods like optical flow \cite{chan2021basicvsr, chan2022basicvsr++, li2022unidirectional}, kernel prediction network \cite{mildenhall2018burst} and spatial-temporal shift \cite{li2023simple} are also employed for achieving temporal alignment. 

\noindent\textbf{Unsupervised Video Denoising.}
The performance of supervised video denoisers is limited by the scarcity of noisy-clean video pairs. To bypass the need for massive paired videos, several studies introduce unsupervised video denoising methods that train solely on noisy videos.
Among these, UDVD~\cite{sheth2021unsupervised} proposes an unsupervised video denoiser utilizing the blind-spot network \cite{krull2019noise2void, laine2019high}. 
RDRF~\cite{wang2023recurrent} improves UDVD's architecture by incorporating a denser receptive field design and recurrent denoising strategy. 
Meanwhile, MF2F~\cite{dewil2021self} presents a Noise2Noise-based~\cite{lehtinen2018noise2noise} unsupervised training approach by constructing noisy pairs for unsupervised denoising. 
VER2R \cite{zheng2023unsupervised} improves the Recorrupted-to-Recorrupted scheme~\cite{pang2021recorrupted} for unsupervised video training. 
Instead of training an unsupervised video denoiser exclusively on noisy videos, Restored-from-Restore~\cite{lee2021restore} (RFR) proposes fine-tuning pre-trained video denoisers to target videos using generated pseudo clean videos.
However, the requirement of a pre-trained video denoiser impedes its practical application due to the lack of video pairs for pre-training.

\section{Methodology}
In this section, we describe the proposed unsupervised video denoising framework TAP. We first introduce the architectures of the video denoiser lifted from a pre-trained image denoiser (Sec.~\ref{modified_video_denoiser}), followed by the unsupervised progressive fine-tuning strategy to harness temporal information across noisy frames (Sec.~\ref{progressive_fine-tuning}).

\subsection{Lifting Image Denoisers for Video Denoising}\label{modified_video_denoiser}  
The challenge of extending a pre-trained image denoiser to a video denoiser lies in preserving the learned spatial information of the pre-trained image denoiser while enabling it to exploit temporal information across multiple noisy frames. 
In this paper, we choose the widely-used encoder-decoder based image denoisers~\cite{cheng2021nbnet, liang2021swinir, wang2022uformer, zamir2022restormer} as the baseline. We then propose to \bihan{integrate} temporal modules into each skip connection between the encoder and decoder of the image denoiser, enforcing the network to effectively learn temporal alignment at various feature levels.
The overall architectures of the video denoiser and details of temporal modules are introduced in the following subsections.

\begin{figure}[tp]
    \centering
    \includegraphics[width=\linewidth]{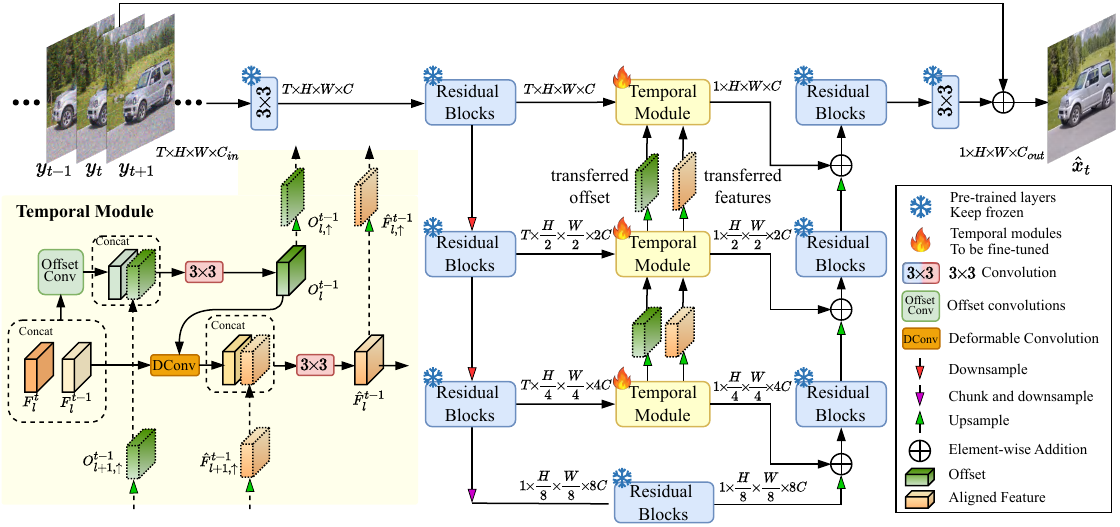}
    \caption{Architecture of the video denoiser. We lift an encoder-decoder based image denoiser (blue part) for video denoising by plugging some temporal modules (yellow part) into its skip connection between the encoder and decoder. Note that parameters from the image denoiser are frozen and the tunable part is only the temporal modules.}
    \label{fig_baseline}
\end{figure}

\noindent\textbf{Overall Pipeline.}
Given a noisy video with $N$ frames $\mathcal{Y} = \{y_1, ...y_t, ...,y_N\}$, our target is to restore the underlying video $\hat{\mathcal{X}} = \{\hat{x}_1,..., \hat{x}_t, ...,\hat{x}_N\}$. 
Our video denoiser takes $T$ noisy frames as input, where $T$ is an odd number, and restores their central frame. 
For sake of clarity, let $Y_t\in\mathbb{R}^{T\times H \times W \times C_{in}}$ denotes the network's input, a sequence of noisy frames clipped from the video $\mathcal{Y}$ and centered at frame $y_t\in\mathbb{R}^{1\times H \times W \times C_{in}}$. 
Let $\hat{x}_t\in\mathbb{R}^{1\times H \times W \times C_{out}}$ denotes the restored central frame. Here, $H$, $W$, $C_{in}$, and $C_{out}$ are the height, width, and number of input and output channels, respectively.
\cref{fig_baseline} illustrates the architecture of our proposed video denoiser, which is extended from a 4-level encoder-decoder based image denoiser.
Each level of encoder-decoder in the image denoiser comprises several residual blocks~\cite{he2016deep} to extract spatial information of each noisy frame. 

To exploit temporal information across noisy frames, temporal modules \cite{wang2019edvr, yue2020supervised} are integrated at skip connections between each level of encoder-decoder.
Except for the bottom level (level-4), features extracted from each noisy frame pass through residual blocks of the encoder. At level-4, only the central frame's features are fed into the residual blocks. 
Temporal modules at each skip connection learn to align deep features of adjacent frames to those of the central frame. 
After alignment, features are fused and added to the same level's reconstruction features of the central frame recovered by the decoder. 

\noindent\textbf{Temporal Module.}
We insert three temporal modules at skip connections from level-1 to level-3, respectively. At each level, temporal modules take deep features $F_l\in \mathbb{R}^{T\times H^\prime \times W^\prime \times C^\prime}$ as input, a concatenation of each frame's deep features $F_l^{t-m}\in \mathbb{R}^{1\times H^\prime \times W^\prime \times C^\prime}$, and output fused features $\hat{F}_l\in \mathbb{R}^{1\times H^\prime \times W^\prime \times C^\prime}$. Here $m$ indicates the time index of noisy frames, which starts from $-T//2$ to $T//2$. $H^\prime$, $W^\prime$, and $C^\prime$ are height, width, and number of channels at level-$l$. More specifically, each temporal module first aligns deep features of neighboring frames with those of the central frame by applying deformable convolution (DCN) \cite{dai2017deformable}. Given deep features of a neighboring frame $F_l^{t-m}$ and the central frame $F_l^t$ at level-$l$, the temporal modules adopt several convolutional layers to learn offsets:
\begin{equation}
    \tilde{O}_l^{t-m} = \mathcal{C}\left( \text{Concat}\left[F_l^t, F_l^{t-m}\right]\right),
\end{equation}
where $\tilde{O}_l^{t-m}$ is the learned offset for deep features $F_l^{t-m}$, $\mathcal{C}$ is the stack of multiple $3\times 3$ convolution layers, and $\text{Concat}\left[\cdot, \cdot \right]$ represents the operation of concatenating along the channel dimension. After that, the learned offset is refined with the offset transferred from the temporal module at the lower level by a $3\times 3$ convolution:
\begin{equation}
    O_l^{t-m} = \text{Conv}\left(\text{Concat}\left[\tilde{O}_l^{t-m}, O_{l+1, \uparrow}^{t-m}\right]\right).
\end{equation}
Here we denote $ O_l^{t-m}$ the refined offset and $O_{l+1, \uparrow}^{t-m}$ the $2 \times$ bilinear upsampled transferred offset from level-$(l\!+\!1)$. Then we apply DCN to align adjacent features $F_l^{t-m}$ and refine them with the transferred features from the lower level. An extra convolution is utilized to fuse these aligned features $\hat{F}_l^{t-m}$: 
\begin{equation}
    \hat{F}_l^{t-m} = \text{Conv}\left( \text{Concat}\left[\mathcal{D}\left(F_l^{t-m}; O_l^{t-m} \right), \hat{F}_{l+1, \uparrow}^{t-m} \right]\right),
\end{equation}
\begin{equation}
    \tilde{F}_{l} = \text{Conv}\left(\text{Concat}\left[\hat{F}_l^{t-T//2},...,\hat{F}_l^t,...,\hat{F}_l^{t+T//2} \right] \right),
\end{equation}
where $\hat{F}_{l-1, \uparrow}^{t-m}$ denotes the transferred aligned features from level-$(l\!+\!1)$, $\tilde{F}_{l}\in\mathbb{R}^{1\times H^\prime \times W^\prime \times C^\prime}$ denotes the fused features, and $\mathcal{D}$ denotes the DCN. Note that the temporal module at the bottom level (level-3) doesn't require the transferred offsets and features for alignment. To this end, the output of temporal module $\hat{F}_l$ is derived as:
\begin{equation}
    \hat{F}_l = F_l^t + \beta \tilde{F}_l,
\end{equation}
where $\beta \in \mathbb{R}^{1\times 1\times 1\times C^\prime}$ is the tunable parameters jointly trained with the temporal module. We initialize $\beta$ to $0$, indicating the network disregards temporal information at the beginning. Consequently, the video denoiser becomes the pre-trained image denoiser. 
This design maximally harnesses the learned spatial information from the original image denoiser while enabling the network to learn temporal information.

\subsection{Unsupervised Progressive Fine-tuning}
\label{progressive_fine-tuning}
\bihan{Given an image denoiser that is trained on a corpus of noisy-clean image pairs, one can directly apply it to denoise the noisy video in a frame-by-frame manner~\cite{wen2018vidosat,zhang2017beyond}. 
Such an approach largely ignores the inter-frame temporal correlation of the video data, resulting in unsatisfied results~\cite{wen2018vidosat}.
To complement the spatial prior, exploiting the temporal information within noisy video data is critical for achieving highly effective video denoising~\cite{wen2018vidosat,tassano2019dvdnet}.}
To this end, we further propose an unsupervised fine-tuning strategy to progressively train the plugged temporal modules solely on collected noisy videos. 

The proposed strategy comprises three steps that progressively fine-tune modules from level-3 to level-1.
\begin{figure}[tp]
    \centering
    \includegraphics[width=.9\linewidth]{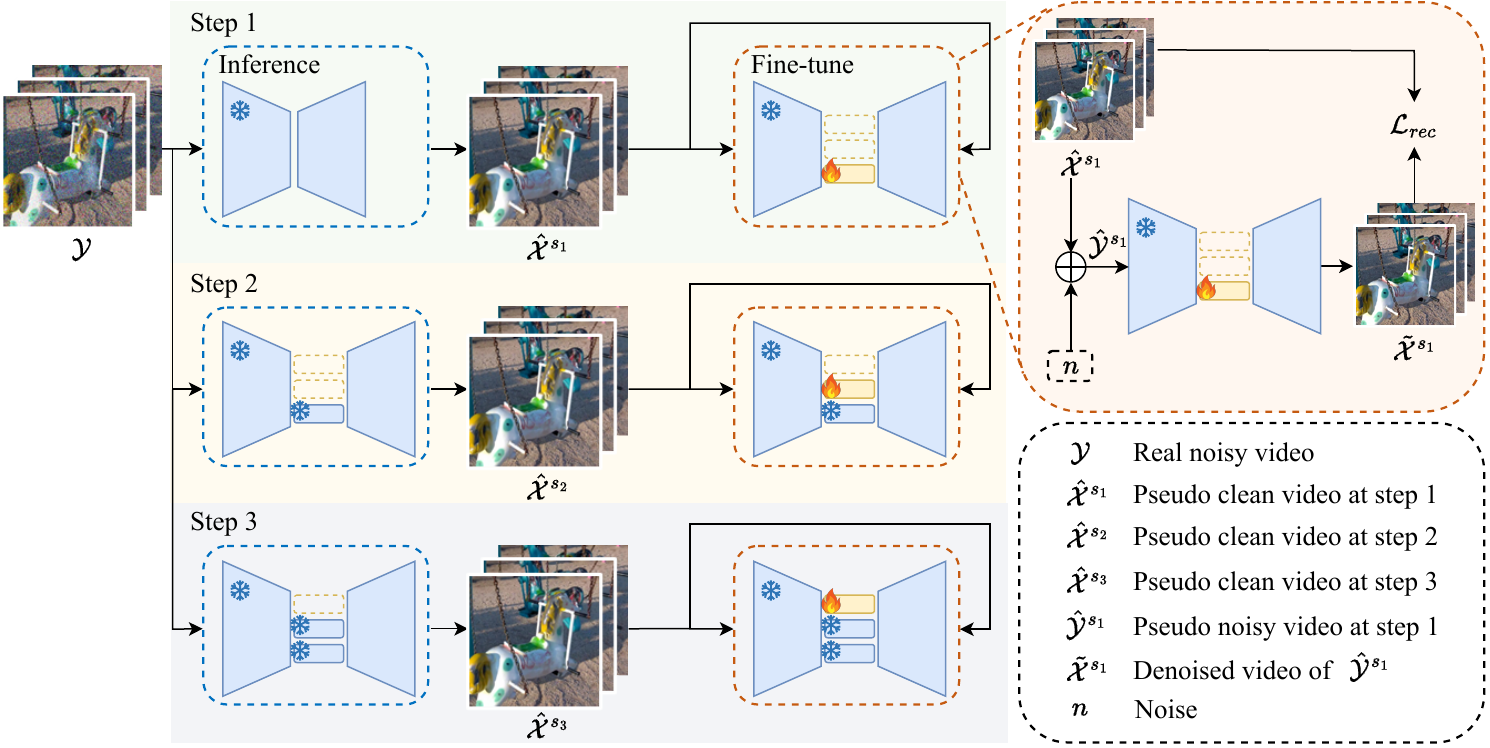}
    \caption{Illustration of proposed unsupervised progressive fine-tuning strategy. The process begins with training the temporal module at level-3, subsequently progressing to the temporal modules at upper levels with pseudo video pairs.}
    \label{fig_frogressive_finetune}
\end{figure}
Let $\mathcal{R}_0$ and $\mathcal{R}_1$ represent the image denoiser and video denoiser, \bihan{respectively}. $\Theta^*$ denotes the parameters of pre-trained image denoiser and $\mathcal{W}_{1:3}$ denotes the parameters of three untrained temporal modules from level-3 to level-1 of the encoder-decoder stages.
For example, $\mathcal{W}_1$ corresponds to the parameters of the temporal module at level-3.
For each inserted temporal module, we construct the noisy and pseudo-clean video pairs to fine-tune the trainable module parameters. 
The illustration of the fine-tuning strategy is described in \cref{fig_frogressive_finetune}. Specifically, given a noisy video dataset $\mathcal{Y}$, the pre-trained image denoiser first processes noisy videos frame by frame:
\begin{equation}
    \hat{x}_t^{s_1} = \mathcal{R}_0(y_t; \Theta^*),
\end{equation}
where $\hat{x}_t^{s_1}$ represents corresponding denoised counterpart of noisy frame $y_t$. Then we construct the pseudo noisy-clean video pairs $\{\hat{\mathcal{Y}}^{s_1}, \hat{\mathcal{X}}^{s_1}\}$, where 
$\hat{\mathcal{X}}^{s_1}$ is the denoised video and $\hat{\mathcal{Y}}^{s_1}$ is the recorrupted video by adding the noise from a known noise model to the denoised video $\hat{\mathcal{X}}^{s_1}$. We fine-tune the temporal module at level-3 on the pseudo noisy-clean video pairs:
\begin{equation}
    \mathcal{W}_1^* = \argmin_{\mathcal{W}_1} \sum_{t=1}^N\lVert \mathcal{R}_1(\hat{Y}_t^{s_1};\Theta^*, \mathcal{W}_1 )-\hat{x}_t^{s_1} \rVert_1,
\end{equation}
where $\hat{Y}_t^{s_1}$ is a contiguous noisy sequence centered at frame $\hat{y}_t^{s_1}$ within the pseudo noisy videos $\hat{\mathcal{Y}}^{s_1}$ and $\mathcal{W}_1^*$ denotes the optimized parameters of $\mathcal{W}_1$. During fine-tuning, instead of updating all parameters of the denoiser, we only update the parameters of the temporal module at the current level. The reason behind this is that denoised video $\hat{\mathcal{X}}^{s_1}$ suffers from over-smoothing, making it unreliable as ground truth. Updating all parameters could potentially undermine the well-learned spatial information of the pre-trained model, resulting in the degradation of network performance. Instead, by solely updating the parameters of the temporal module, we can maintain the well-learned spatial information while exploiting temporal information.
In subsequent steps, we follow the similar training process as previous by constructing new pseudo noisy-clean video pairs $\{ \hat{\mathcal{Y}}^{s_m}, \hat{\mathcal{X}}^{s_m}\}, m=2,3$, and progressively fine-tuning the plugged temporal modules at level-2 and level-1. The only difference is that the pseudo clean video $\hat{\mathcal{X}}^{s_m}$ are obtained by applying the fine-tuned video denoiser in the previous step:
\begin{equation}
    \hat{x}_t^{s_m} = \mathcal{R}_1(Y_t; \Theta^*, \mathcal{W}_{1:m-1}^*),
\end{equation}
where $\hat{x}_t^{s_m}$ denotes a denoised frame in the video $\hat{\mathcal{X}}^{s_m}$ at step $m$. 
Subsequently, parameters of temporal modules at level-2 and level-1 are optimized as:
\begin{equation}
    \mathcal{W}_m^* = \argmin_{\mathcal{W}_m} \sum_{t=1}^N\lVert \mathcal{R}_1(\hat{Y}_t^{s_m};\Theta^*, \mathcal{W}_{1:m-1}^*, \mathcal{W}_m )-\hat{x}_t^{s_m} \rVert_1.
\end{equation}
Once all temporal modules are well-trained, the network can be effectively deployed for processing captured noisy videos:
\begin{equation}
    \hat{x}_t = \mathcal{R}_1(Y_t;\Theta^*, \mathcal{W}_{1:m}^*).
\end{equation}

\section{Experiments}
\subsection{Experimental Settings}
\subsubsection{Datasets.} 
We demonstrate the effectiveness of our proposed method on two denoising tasks: synthetic Gaussian video denoising and real raw video denoising. 
We use DAVIS~\cite{pont20172017} and Set8~\cite{tassano2020fastdvdnet} for synthetic Gaussian video denoising and CRVD~\cite{yue2020supervised} for real raw video denoising. 
DAVIS includes 90 videos for training and validation and 30 videos for testing while Set8 consists of 4 videos from Derf's Test Media collection and 4 videos shot with GoPro, exclusively utilized for evaluation.
CRVD provides indoor and outdoor raw videos. 
The indoor set includes 11 scenes of paired noisy-clean videos (6 scenes for training and 5 scenes for testing), with each scene consisting of 7 video frames.
The outdoor set only includes noisy videos captured in complex scenes with realistic motion, mainly used for visual comparison.

\noindent\textbf{Implementation Details.} 
We separately train image denoisers on synthetic Gaussian image denoising datasets and real raw image denoising datasets. 
We use an encoder-decoder based model, \ie, NAFNet~\cite{chen2022simple}, as the baseline image denoiser for all experiments.
Following previous works~\cite{zamir2022restormer, zhang2021plug}, we use a combination of BSD500~\cite{martin2001database}, DIV2K~\cite{agustsson2017ntire}, Flickr2K~\cite{lim2017enhanced}, and WED~\cite{ma2016waterloo} to train the Gaussian image denoiser. Noisy images are synthesized by adding additive white Gaussian noise (AWGN) of $\sigma\in [10, 55]$ to clean images. 
To train the raw image denoiser, we select clean images from SID~\cite{chen2018learning}, SIDD~\cite{abdelhamed2018high}, and the CRVD training set. 
We synthesize Poisson-Gaussian noise on these clean images using camera noise model parameters provided in CRVD. Note that CRVD is a video dataset, we split videos into individual images for training. 
To ensure the fairness of our experiments, the CRVD training set is not used to fine-tune the plugged temporal modules. 
For our video denoiser, the input time frame number $T$ is set to 5 for Gaussian denoising and 3 for real raw video denoising. The patch size of training sequences is randomly cropped to $160\times 160$ and the batch size is 4. Raw videos are packed into RGBG sequences during training. We fine-tune the network with Adam optimizer~\cite{kingma2014adam} and $\mathcal{L}_1$ loss. At each fine-tuning step, the network is trained for 200$K$ iterations, where the learning rate is initially set to $1e^{-3}$ and gradually decreases to $1e^{-5}$ with cosine annealing~\cite{loshchilov2016sgdr}. The method is implemented with PyTorch and conducted on one NVIDIA RTX A5000. More detailed architectures can be found in the \textbf{supplementary material}.

\subsection{Comparison with State-of-the-Art}
We compare our proposed TAP with state-of-the-art unsupervised video denoising methods, \ie, MF2F~\cite{dewil2021self}, RFR~\cite{lee2021restore}, UDVD~\cite{sheth2021unsupervised}, RDRF~\cite{wang2023recurrent}, and ER2R~\cite{zheng2023unsupervised}. 
Unsupervised methods, not reliant on ground truth, can be directly trained and evaluated on the test set or trained on the training set before evaluation on the test set. 
For synthetic Gaussian denoising, MF2F, RFR, and ER2R are directly trained and assessed on the Set8 and DAVIS test set, while UDVD and RDRF are trained using the DAVIS training set and evaluated on the test set. To differentiate between the two settings, we tag `T' to methods trained directly on the test set, \eg, ER2R-T. We trained TAP and TAP-T for these two settings. 
For real raw video denoising, due to the limited size of CRVD, all unsupervised methods are trained on the CRVD indoor test set and CRVD outdoor test set. The experimental results from the competing methods are either directly cited from the original papers or reproduced by running the official code. In cases where the code is unavailable, we omit the corresponding results. 
For a comprehensive comparison, traditional and supervised video denoising methods such as VBM4D~\cite{maggioni2012video}, FastDVDNet~\cite{tassano2020fastdvdnet}, PaCNet~\cite{vaksman2021patch}, RViDeNet~\cite{yue2020supervised}, and FloRNN~\cite{li2022unidirectional} are also included. 
Additionally, we report the performance of our pre-trained image denoiser on video denoising, \ie, NAFNet~\cite{chen2022simple}. 
%

\begin{figure}[!t]
    \centering
    \includegraphics[width=\linewidth]{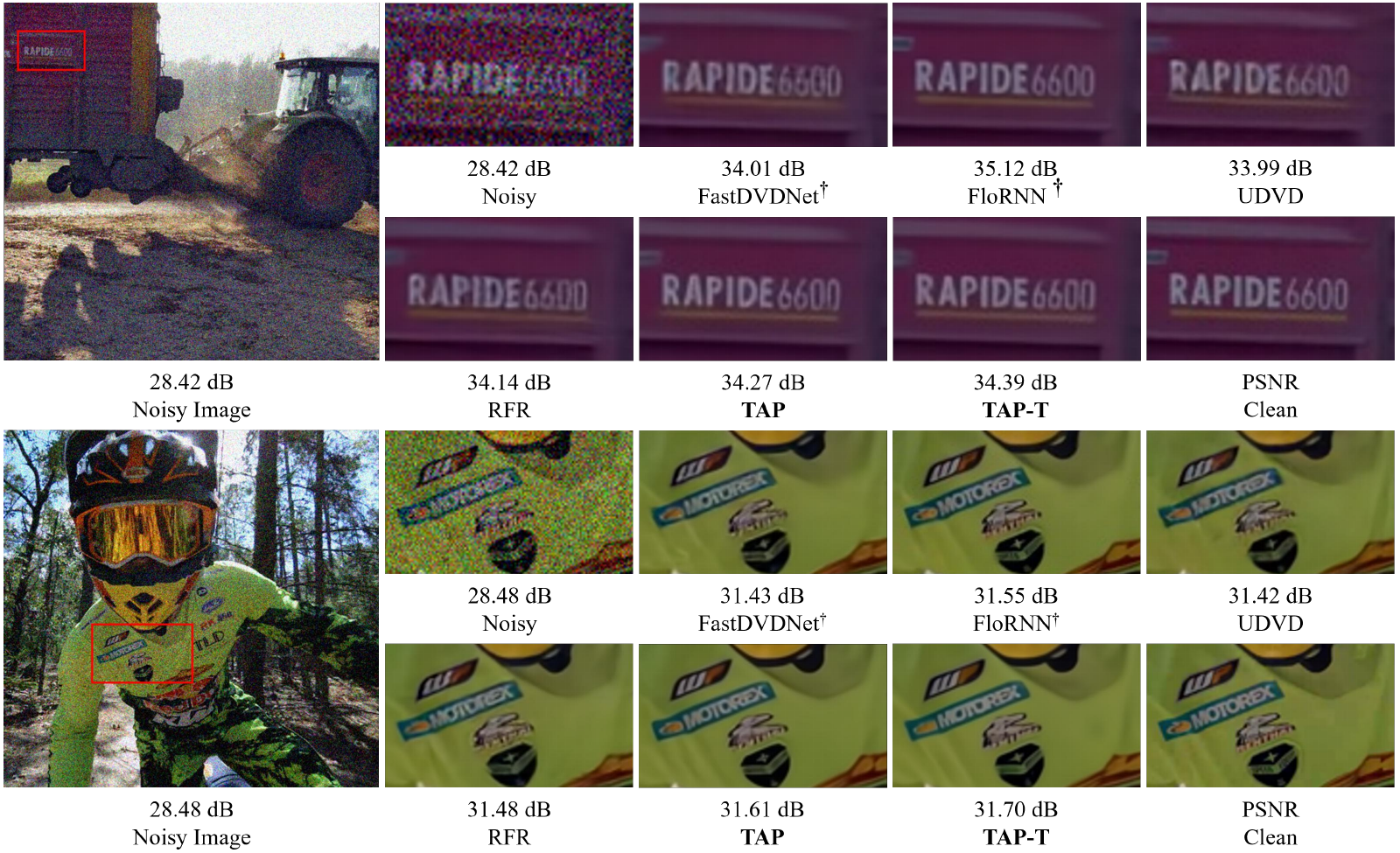}
    \caption{Visual examples of the synthetic Gaussian denoising results on DAVIS (top) and Set8 (bottom) datasets, including the noisy inputs (with noise level $\sigma=30$), the restored images using FastDVDNet~\cite{tassano2020fastdvdnet}, FloRNN~\cite{li2022unidirectional}, UDVD~\cite{sheth2021unsupervised}, RFR~\cite{lee2021restore}, TAP, and TAP-T, as well as the clean images, respectively. \dag~indicates the supervised method.}
    \label{fig_davis_set8}
\end{figure}

\noindent\textbf{Experiments on synthetic Gaussian video denoising.} 
\cref{tab1_davis} presents the quantitative video denoising results on the DAVIS and Set8 datasets. It is obvious that our method outperforms other unsupervised methods by large margins. Notably, TAP-T increases PSNR from 32.92 dB to 33.95 dB compared to RDRF over the Set8 dataset and also outperforms the supervised method FloRNN~\cite{li2022unidirectional} on PSNR. This improvement is partly attributed to the relatively smooth motion in the Set8 dataset, which facilitates the network to effectively learn temporal information by directly fine-tuning over the test set.  In the meantime, TAP achieves second denoising performance in PSNR among unsupervised denoisers. Compared to our pre-trained image denoiser, TAP-T gains 1.24 dB and 1.52 dB improvement on the DAVIS and Set8 datasets respectively, suggesting our network effectively learns temporal information from the noisy frames. The results of both TAP and TAP-T demonstrate the effectiveness of our proposed framework.
\cref{fig_davis_set8} illustrates video denoising examples on DAVIS and Set8 datasets. The unsupervised methods, UDVD and RFR, fail to recover fine details, resulting in visible artifacts. In contrast, our methods, TAP and TAP-T, effectively restore details (evident in the clarity of lines and textures in the denoised images) and achieve comparable results to that of the supervised method FloRNN. More visual examples are provided in the \textbf{supplementary material}. 

\begin{table*}[!t]
    \centering
        \caption{Quantitative evaluation (using PSNR and SSIM) of video denoising results on the DAVIS~\cite{pont20172017} and Set8~\cite{tassano2020fastdvdnet} datasets, using our TAP framework, comparing to the representative methods that are traditional, supervised or unsupervised approaches. The best and second results among unsupervised methods are \textbf{highlighted} and \underline{underline}.}
    \adjustbox{width=1.\linewidth}{
    \begin{tabular}{c|c|l|cccccccccccc}
    \toprule
    \multirow{2}{*}{} & \multirow{2}{*}{}& \multirow{2}{*}{Method} & \multicolumn{2}{c}{$\sigma=10$} & \multicolumn{2}{c}{$\sigma=20$} & \multicolumn{2}{c}{$\sigma=30$} & \multicolumn{2}{c}{$\sigma=40$} & \multicolumn{2}{c}{$\sigma=50$} & \multicolumn{2}{c}{Average} \\
    & & & PSNR & SSIM & PSNR & SSIM & PSNR & SSIM & PSNR & SSIM & PSNR & SSIM & PSNR & SSIM\\
    \midrule
    \multirow{10}{*}{\rotatebox{90}{DAVIS}} & Traditional & VBM4D~\cite{maggioni2012video} & 37.58 & - & 33.88 & - & 31.65 & - & 30.05 & - & 28.80 & - & 32.39 & - \\ \cline{2-15}
    & \multirow{4}{*}{Supervised} & NAFNet~\cite{chen2022simple} & 38.79 & 0.965 & 35.37 & 0.933 & 33.47 & 0.904 & 32.17 & 0.879 & 31.18 & 0.858 & 34.20 & 0.908 \\
    & & FastDVDNet~\cite{tassano2020fastdvdnet} & 38.71 & 0.962 & 35.77 & 0.941 & 34.04 & 0.917 & 32.82 & 0.895 & 31.86 & 0.875 & 34.64 & 0.919 \\
    & & PaCNet~\cite{vaksman2021patch} & 39.97 & 0.971 & 37.10 & 0.947 & 35.07 & 0.921 & 33.57 & 0.897 & 32.39 & 0.874 & 35.62 & 0.922 \\
    & & FloRNN~\cite{li2022unidirectional} & 40.16 & 0.976 & 37.52 & 0.956 & 35.89 & 0.944 & 34.66 & 0.929 & 33.67 & 0.913 & 36.38 & 0.944 \\ \cline{2-15}
    & \multirow{6}{*}{Unsupervised} & MF2F-T~\cite{dewil2021self} & 38.04 & 0.957 & 35.61 & 0.936 & 33.65 & 0.907 & 31.50 & 0.852 & 29.39 & 0.784 & 33.64 & 0.887 \\
    & & RFR-T~\cite{lee2021restore} & 39.31 & - & 36.15 & - & 34.28 & - & 32.92 & - & 31.86 & - & 34.90 & - \\
    & & UDVD~\cite{sheth2021unsupervised} & 39.17 & 0.970 & 35.94 & 0.943 & 34.09 & 0.918 & 32.79 & 0.895 & 31.80 & 0.874 & 34.76 & 0.920 \\
    & & RDRF~\cite{wang2023recurrent} & 39.54 & \underline{0.972} & 36.40 & 0.947 & 34.55 & \underline{0.925} & 33.23 & \underline{0.903} & 32.20 & 0.883 & 35.18 & \underline{0.926} \\
    & & ER2R-T~\cite{zheng2023unsupervised} & 39.52 & - & 36.49 & - & 34.60 & - & 33.29 & - & 32.25 & - & 35.23 & - \\
    & & \cellcolor{Gray}TAP & \cellcolor{Gray}\underline{39.69} & \cellcolor{Gray}\underline{0.972} & \cellcolor{Gray}\underline{36.62} & \cellcolor{Gray}\underline{0.948} & \cellcolor{Gray}\underline{34.71} & \cellcolor{Gray}\underline{0.925} &  \cellcolor{Gray} \underline{33.37} & \cellcolor{Gray}\underline{0.903} & \cellcolor{Gray}\underline{32.36} & \cellcolor{Gray}\underline{0.884} & \cellcolor{Gray}\underline{35.35} & \cellcolor{Gray}\underline{0.926}\\ 
    & & \cellcolor{Gray}TAP-T & \cellcolor{Gray}\textbf{39.80} & \cellcolor{Gray}\textbf{0.973} & \cellcolor{Gray}\textbf{36.74} & \cellcolor{Gray}\textbf{0.950} & \cellcolor{Gray}\textbf{34.84} & \cellcolor{Gray}\textbf{0.926} &  \cellcolor{Gray}\textbf{33.49} & \cellcolor{Gray}\textbf{0.905} & \cellcolor{Gray}\textbf{32.49} & \cellcolor{Gray}\textbf{0.886} & \cellcolor{Gray}\textbf{35.48} & \cellcolor{Gray}\textbf{0.928}\\ 

    \midrule
    \multirow{10}{*}{\rotatebox{90}{Set8}} & Traditional & VBM4D~\cite{maggioni2012video} & 36.05 & - & 32.19 & - & 30.00 & - & 28.48 & - & 27.33 & - & 30.81 & - \\ \cline{2-15}
    & \multirow{4}{*}{Supervised} & NAFNet~\cite{chen2022simple} & 36.52 & 0.943 & 33.55 & 0.802 & 31.81 & 0.869 & 30.59 & 0.842 & 29.65 & 0.818 & 32.43 & 0.875 \\
    & & FastDVDNet~\cite{tassano2020fastdvdnet} & 36.44 & 0.954 & 33.43 & 0.920 & 31.68 & 0.889 & 30.46 & 0.861 & 29.53 & 0.835 & 32.31 & 0.892 \\
    & & PaCNet~\cite{vaksman2021patch} & 37.06 & 0.960 & 33.94 & 0.925 & 32.05 & 0.892 & 30.70 & 0.862 & 29.66 & 0.835 & 32.68 & 0.895 \\
    & & FloRNN~\cite{li2022unidirectional} & 37.57 & 0.964 & 34.67 & 0.938 & 32.97 & 0.914 & 31.75 & 0.891 & 30.80 & 0.870 & 33.55 & 0.915 \\ \cline{2-15}
    & \multirow{6}{*}{Unsupervised} & MF2F-T~\cite{dewil2021self} & 36.01 & 0.938 & 33.79 & 0.912 & 32.20 & 0.883 & 30.64 & 0.841 & 28.90 & 0.778 & 32.31 & 0.870 \\
    & & RFR-T~\cite{lee2021restore} & 36.77 & - & 33.64 & - & 31.82 & - & 30.52 & - & 29.50 & - & 32.45 & - \\
    & & UDVD~\cite{sheth2021unsupervised} & 36.36 & 0.951 & 33.53 & 0.917 & 31.88 & 0.887 & 30.72 & 0.859 & 29.81 & 0.835 & 32.46 & 0.890 \\
    & & RDRF~\cite{wang2023recurrent} & 36.67 & 0.955 & 34.00 & \underline{0.925} & 32.39 & 0.898 & \underline{31.23} & \underline{0.873} & \underline{30.31} & 0.850 & 32.92 & 0.900 \\
    & & ER2R-T~\cite{zheng2023unsupervised} & 37.55 & - & 34.34 & - & 32.45 & - & 31.09 & - & 30.05 & - & 33.10 & - \\
    & & \cellcolor{Gray}TAP & \cellcolor{Gray}\underline{37.76} & \cellcolor{Gray}\underline{0.956} & \cellcolor{Gray}\underline{34.51} & \cellcolor{Gray}\underline{0.925} & \cellcolor{Gray}\underline{32.76} & \cellcolor{Gray}\underline{0.899} & \cellcolor{Gray}{31.21} & \cellcolor{Gray}\underline{0.873} & \cellcolor{Gray}{30.27} & \cellcolor{Gray}\underline{0.851} & \cellcolor{Gray}\underline{33.30} & \cellcolor{Gray}\underline{0.901}\\ 
    & & \cellcolor{Gray}TAP-T & \cellcolor{Gray}\textbf{38.02} & \cellcolor{Gray}\textbf{0.958} & \cellcolor{Gray}\textbf{35.07} & \cellcolor{Gray}\textbf{0.927} & \cellcolor{Gray}\textbf{33.42} & \cellcolor{Gray}\textbf{0.900} &  \cellcolor{Gray}\textbf{32.10} & \cellcolor{Gray}\textbf{0.875} & \cellcolor{Gray}\textbf{31.16} & \cellcolor{Gray}\textbf{0.852} & \cellcolor{Gray}\textbf{33.95} & \cellcolor{Gray}\textbf{0.902}\\ 
    
    \bottomrule
    \end{tabular}}
    \label{tab1_davis}
\end{table*}

\begin{figure}[!t]
    \centering
    \includegraphics[width=\linewidth]{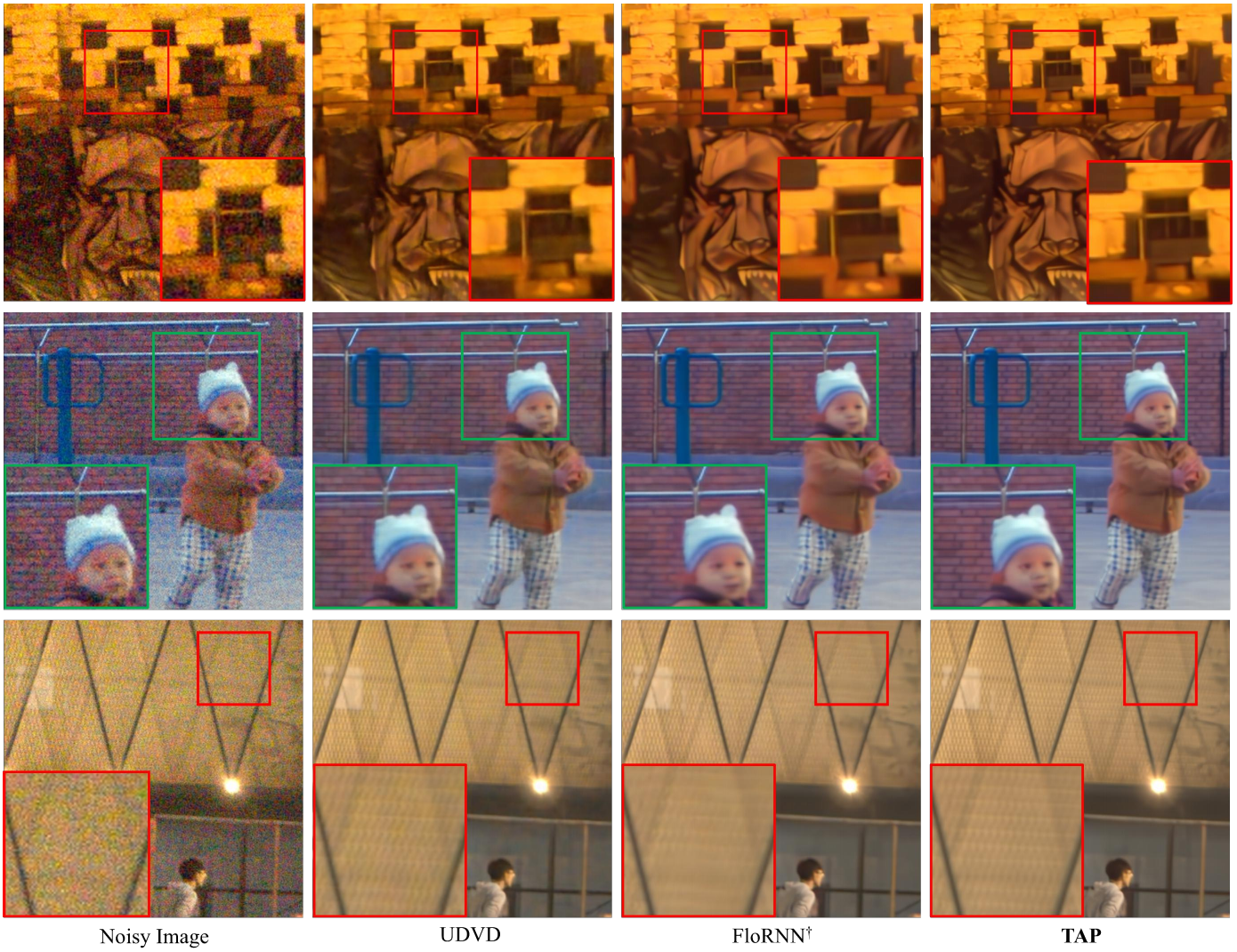}
    \caption{Visual examples of real raw video denoising on CRVD outdoor set. The noisy image, the restored images of UDVD~\cite{sheth2021unsupervised}, FloRNN~\cite{li2022unidirectional}, and TAP, respectively. \dag~denotes the supervised method. We render raw images to sRGB images with the pre-trained ISP provided in~\cite{yue2020supervised} (ISO$=$25600).}
    \label{fig_crvd_outdoor}
\end{figure}

\noindent\textbf{Experiments on real raw video denoising.} We evaluate our method on the CRVD indoor test set and outdoor test set. \cref{tab2_CRVD} shows the quantitative results on the CRVD indoor test set, where our method still outperforms other unsupervised counterparts and presents similar results as FloRNN on PSNR. However, the improvement on the CRVD indoor set is less significant compared to synthetic Gaussian denoising datasets, likely due to the limited size and manually simulated motion of the indoor dataset.
We further demonstrate our method on the outdoor set which provides videos with realistic motion and complex scenes. \cref{fig_crvd_outdoor} illustrates the visual quality of denoised results. FloRNN tends to generate over-smooth results, indicating its difficulty in handling unseen scenes as training video pairs are limited and simulated. Unsupervised denoiser UDVD cannot effectively suppress the noise. Among these methods, our method efficiently eliminates noise while preserving fine details.

\begin{table}[!t]
    \centering
    \caption{Quantitative evaluation (PSNR in dB) of raw video denoising results on the CRVD~\cite{yue2020supervised} indoor test set, using our TAP framework, comparing to the recent raw video denoising methods. The best results among unsupervised methods are \textbf{highlighted}.}
      \adjustbox{width=.75\linewidth}{
    \begin{tabular}{c|l|cccccc}
    \toprule
    \addlinespace[0pt]
     & \diagbox[dir=NW]{Method}{\raisebox{-1.8ex}[0pt]{ISO}} & 1600 & 3200 & 6400 & 12800 & 25600 & Average \\

     \cline{1-8}

    \multirow{3}{*}{Supervised} & NAFNet~\cite{chen2022simple} & 48.02 & 46.05 & 44.04 & 41.44 & 41.49 & 44.21  \\
    & RViDeNet~\cite{yue2020supervised} & 47.74 & 45.91 & 43.85 & 41.20 & 41.17 & 43.97\\
    & FloRNN~\cite{li2022unidirectional} & 48.81 & 47.05 & 45.09 & 42.63 & 42.19 & 45.15 \\ \cline{1-8}
    \multirow{3}{*}{Unsupervised} & UDVD~\cite{sheth2021unsupervised} & 48.02 & 46.44 & 44.74 & 42.21 & 42.13 & 44.71 \\
    & RDRF~\cite{wang2023recurrent} & 48.38 & 46.86 &\textbf{ 45.24} & \textbf{42.72} & 42.25 & 45.09 \\
    & \cellcolor{Gray}Ours & \cellcolor{Gray}{\textbf{48.85}} & \cellcolor{Gray}{\textbf{47.03}} & \cellcolor{Gray}{45.11} &  \cellcolor{Gray}{42.44} & \cellcolor{Gray}{\textbf{42.33}} & \cellcolor{Gray}{\textbf{45.15}} \\ 
    \bottomrule
    \end{tabular}}
\label{tab2_CRVD}
\end{table}

\subsection{Visualization and Analysis of Progressive Fine-Tuning}
\begin{figure}[!t]
    \centering
    \includegraphics[width=\linewidth]{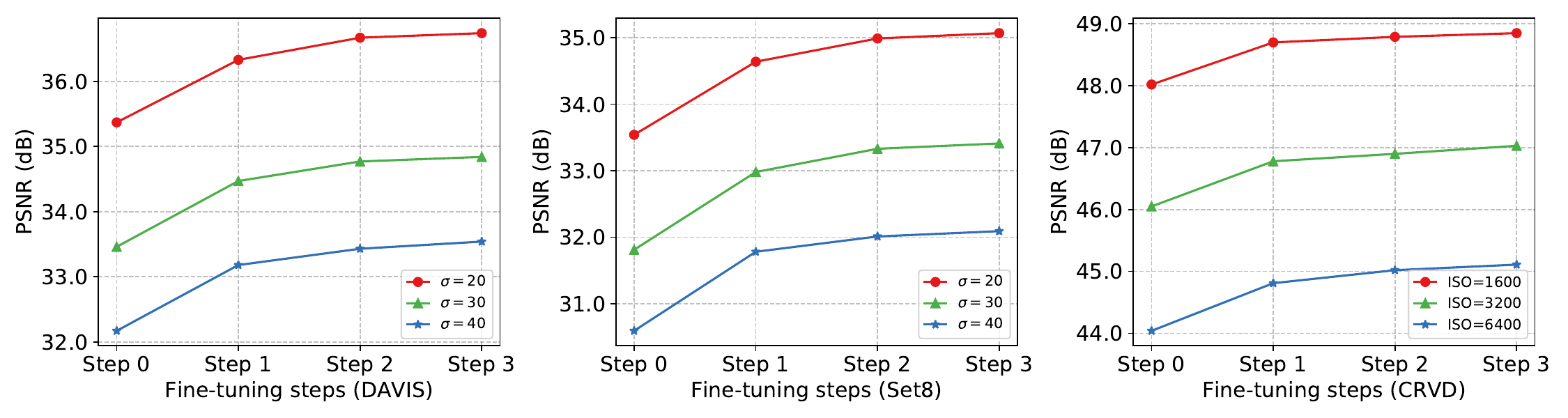}
    \caption{Evaluation of the effectiveness of progressive fine-tuning strategy. The quantitative results are demonstrated on DAVIS, Set8, and CRVD datasets respectively. Step 0 indicates the pre-trained image denoiser.}
    \label{fig_effectiveness}
\end{figure}

\begin{figure}[!t]
    \centering
    \includegraphics[width=\linewidth]{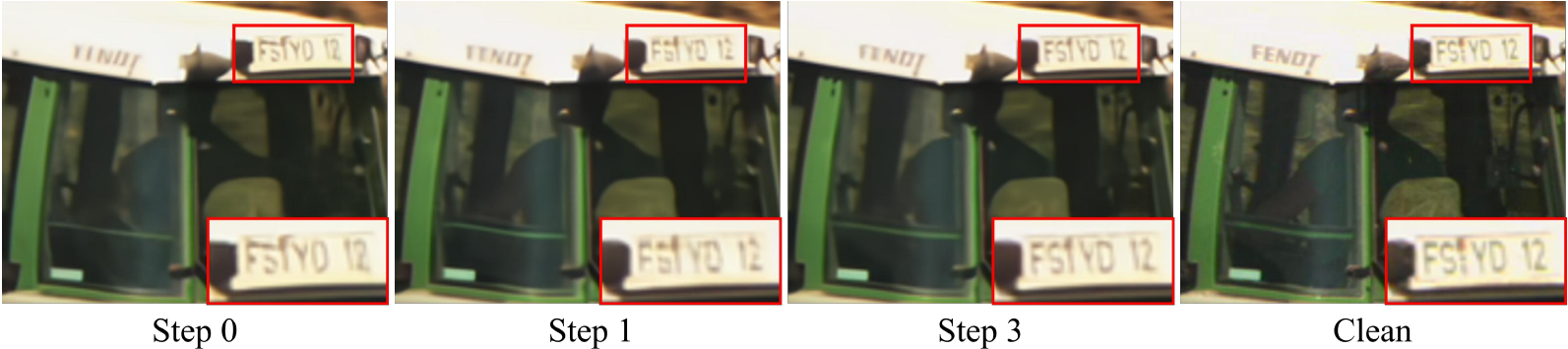}
    \caption{Illustration of progressive fine-tuning strategy (noise level $\sigma=30$).}
    \label{fig_progressive_finetune}
\end{figure}

By incorporating temporal information, the network is supposed to gradually restore details over several fine-tuning steps. 
To demonstrate the effectiveness of our proposed progressive fine-tuning strategy, we conduct experiments on DAVIS, Set8, and CRVD datasets, respectively, and assess networks' denoising performance at each step.
\cref{fig_effectiveness} presents the quantitative results, illustrating a consistent improvement in denoising performance during the fine-tuning process. 
\cref{fig_progressive_finetune} highlights the quality 
enhancement at each step, with the pre-trained denoiser (step 0) producing over-smoothness and artifacts. As training progresses, the network gradually restores detailed texture, demonstrating the method's effectiveness in improving visual quality over steps.

\subsection{Ablation Study}

\begin{table}[!t]
    \centering
    \caption{Ablation on different fine-tuning strategies. Original: fine-tuning temporal modules progressively. Case 1: fine-tuning all parameters progressively. Case 2: fine-tuning three temporal modules jointly. Case 3: repeat the fine-tuning process twice, we report the second time denoising results. Experiments are conducted on the DAVIS dataset with noise level $\sigma=30$.}
    \adjustbox{width=.65\linewidth}{
    \begin{tabular}{c@{\hspace{14pt}}c@{\hspace{14pt}}c@{\hspace{14pt}}c@{\hspace{14pt}}c}
    \toprule
     & Step 0 & Step 1 & Step 2 & Step 3 \\
    \midrule
        Original & 33.47 dB & 34.47 dB & 34.77 dB & 34.84 dB \\
        Case 1 & 33.47 dB & 34.02 dB & 34.14 dB & 34.20 dB \\
        Case 2 & 33.47 dB & 34.49 dB & 34.62 dB & 34.55 dB \\
        Case 3 & - & 34.54 dB & 34.81 dB & 34.86 dB \\
    \bottomrule
    \end{tabular}}
    \label{tab4_ablation_tune}
\end{table}

\noindent\textbf{Fine-tuning all parameters or fine-tuning temporal modules.}
To avoid overfitting to pseudo clean videos, we recommend freezing the pre-trained image denoiser's parameters during fine-tuning. This claim is evaluated by comparing this strategy against a setting where all parameters are fine-tuned, following our progressive fine-tuning strategy at each step. \cref{tab4_ablation_tune} Case 1 presents the quantitative comparisons between fine-tuning all parameters and fine-tuning only the temporal modules. The results show the denoising performance decreases when all parameters are updated, likely due to overfitting. Thus we opt to freeze pre-trained layers during fine-tuning.

\noindent\textbf{Fine-tuning temporal modules jointly or progressively.}
We explore an alternative fine-tuning strategy by simultaneously training all modules. To verify its effectiveness against our proposed strategy, we freeze the parameters of pre-trained layers and jointly fine-tune all temporal modules. For a fair comparison, the fine-tuning process is repeated three times, similar to our method. 
The quantitative results, presented in \cref{tab4_ablation_tune} Case 2, show the denoising performance increases at step one under that configuration. However, its performance decreases in the subsequent fine-tuning steps. This degradation is attributed to network overfitting to the over-smooth results and the instability caused by jointly training all temporal modules, leading to the learned offsets with significant magnitudes. Moreover, training all temporal modules at each step leads to the high computational cost compared to the progressive fine-tuning strategy. 

\noindent\textbf{Fine-tuning with more steps.}
We further explore the effectiveness of repeating the whole fine-tuning process compared to the original strategy. After completing fine-tuning at step 3, we repeat the procedure with the pseudo pairs. According to \cref{tab4_ablation_tune} Case 3, fine-tuning twice yields marginally better results than once   . Therefore, we choose not to repeat the whole fine-tuning process in this study.

\noindent\textbf{Temporal Coherence.} Following previous works \cite{song2022tempformer, li2023simple}, we create a static 12-frame toy dataset with noise level $\sigma=30$. 
\begin{table}[t]
    \centering
    \caption{Quantitative evaluation of temporal coherence on denoised frames. \textbf{Lower is better}.}
    \adjustbox{width=.65\linewidth}{\begin{tabular}{ccccc}
        \hline
         NAFNet$^{\dag}$ & FastDVD$^{\dag}$ & FloRNN$^{\dag}$ & UDVD & \textbf{Ours} \\
        $16.5\times10^{-3}$ & $6.9\times10^{-3}$ & \textbf{$1.7\times10^{-3}$} & $10.9\times10^{-3}$ & \textbf{$5.8\times10^{-3}$} \\
         \hline
    \end{tabular}}
    \label{tab5_temporal_coherence}
\end{table}
\begin{figure}[!t]
    \centering
    \includegraphics[width=\linewidth]{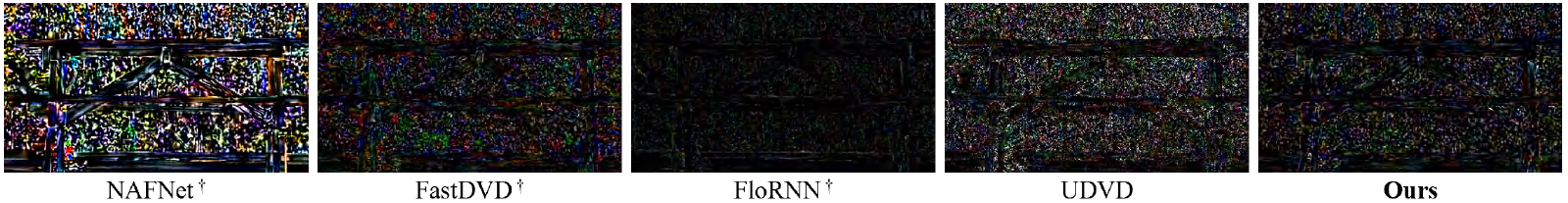}
    \caption{Visualization of a residual image between two adjacent frames on the toy dataset. $\dag$ denotes supervised methods. Better to zoom in.}
    \label{fig_temporal_coherence}
\end{figure}
We calculate the mean absolute error of adjacent denoised frames and visualize one of the residual images between them to evaluate temporal coherence. \cref{tab5_temporal_coherence} and \cref{fig_temporal_coherence} show our denoiser has better temporal coherence than the unsupervised competitor UDVD. Moreover, TAP improves temporal coherence compared to our image denoising baseline NAFNet.

\section{Conclusion}
In this paper, we propose to utilize a pre-trained image denoiser as a spatial prior for unsupervised video denoising. We introduce a novel unsupervised video denoising framework TAP which enables an image denoiser to effectively exploit temporal information exclusively on noisy videos. We extend an image denoiser to a video denoiser by inserting trainable temporal modules and propose a progressively fine-tuning strategy to train each temporal module correspondingly. Finally, we evaluate TAP on synthetic Gaussian video denoising and real raw video denoising tasks, which achieve superior denoising performance compared to recent start-of-the-art methods.

\noindent\textbf{Limitations and Future Work.}
Our framework demonstrates effectiveness but could benefit from sophisticated architectural design. Integrating optical flow~\cite{sun2018pwc, xue2019video} alongside deformable convolution could facilitate temporal alignment. Meanwhile, incorporating attention mechanism~\cite{wang2019edvr, yue2020supervised} has the potential to further improve denoising results. These approaches suggest further exploration of the proposed framework.

\section*{Acknowledgments}
This work was partially supported by the National Research Foundation Singapore Competitive Research Program (award number CRP29-2022-0003). This research was carried out at the Rapid-Rich Object Search (ROSE) Lab, Nanyang Technological University, Singapore.
%
%
\bibliographystyle{splncs04}
\bibliography{egbib}

\begin{thebibliography}{10}
\providecommand{\url}[1]{\texttt{#1}}
\providecommand{\urlprefix}{URL }
\providecommand{\doi}[1]{https://doi.org/#1}

\bibitem{abdelhamed2018high}
Abdelhamed, A., Lin, S., Brown, M.S.: A high-quality denoising dataset for
  smartphone cameras. In: Proceedings of the IEEE conference on computer vision
  and pattern recognition. pp. 1692--1700 (2018)

\bibitem{agustsson2017ntire}
Agustsson, E., Timofte, R.: Ntire 2017 challenge on single image
  super-resolution: Dataset and study. In: Proceedings of the IEEE conference
  on computer vision and pattern recognition workshops. pp. 126--135 (2017)

\bibitem{chan2021basicvsr}
Chan, K.C., Wang, X., Yu, K., Dong, C., Loy, C.C.: Basicvsr: The search for
  essential components in video super-resolution and beyond. In: Proceedings of
  the IEEE/CVF conference on computer vision and pattern recognition. pp.
  4947--4956 (2021)

\bibitem{chan2022basicvsr++}
Chan, K.C., Zhou, S., Xu, X., Loy, C.C.: Basicvsr++: Improving video
  super-resolution with enhanced propagation and alignment. In: Proceedings of
  the IEEE/CVF conference on computer vision and pattern recognition. pp.
  5972--5981 (2022)

\bibitem{chang2020spatial}
Chang, M., Li, Q., Feng, H., Xu, Z.: Spatial-adaptive network for single image
  denoising. In: Computer Vision--ECCV 2020: 16th European Conference, Glasgow,
  UK, August 23--28, 2020, Proceedings, Part XXX 16. pp. 171--187. Springer
  (2020)

\bibitem{chen2019seeing}
Chen, C., Chen, Q., Do, M.N., Koltun, V.: Seeing motion in the dark. In:
  Proceedings of the IEEE/CVF International conference on computer vision. pp.
  3185--3194 (2019)

\bibitem{chen2018learning}
Chen, C., Chen, Q., Xu, J., Koltun, V.: Learning to see in the dark. In:
  Proceedings of the IEEE conference on computer vision and pattern
  recognition. pp. 3291--3300 (2018)

\bibitem{chen2022simple}
Chen, L., Chu, X., Zhang, X., Sun, J.: Simple baselines for image restoration.
  In: European Conference on Computer Vision. pp. 17--33. Springer (2022)

\bibitem{cheng2021nbnet}
Cheng, S., Wang, Y., Huang, H., Liu, D., Fan, H., Liu, S.: Nbnet: Noise basis
  learning for image denoising with subspace projection. In: Proceedings of the
  IEEE/CVF conference on computer vision and pattern recognition. pp.
  4896--4906 (2021)

\bibitem{dai2017deformable}
Dai, J., Qi, H., Xiong, Y., Li, Y., Zhang, G., Hu, H., Wei, Y.: Deformable
  convolutional networks. In: Proceedings of the IEEE international conference
  on computer vision. pp. 764--773 (2017)

\bibitem{dewil2021self}
Dewil, V., Anger, J., Davy, A., Ehret, T., Facciolo, G., Arias, P.:
  Self-supervised training for blind multi-frame video denoising. In:
  Proceedings of the IEEE/CVF winter conference on applications of computer
  vision. pp. 2724--2734 (2021)

\bibitem{dudhane2023burstormer}
Dudhane, A., Zamir, S.W., Khan, S., Khan, F.S., Yang, M.H.: Burstormer: Burst
  image restoration and enhancement transformer. In: 2023 IEEE/CVF Conference
  on Computer Vision and Pattern Recognition (CVPR). pp. 5703--5712. IEEE
  (2023)

\bibitem{he2016deep}
He, K., Zhang, X., Ren, S., Sun, J.: Deep residual learning for image
  recognition. In: Proceedings of the IEEE conference on computer vision and
  pattern recognition. pp. 770--778 (2016)

\bibitem{ioffe2015batch}
Ioffe, S., Szegedy, C.: Batch normalization: Accelerating deep network training
  by reducing internal covariate shift. In: International conference on machine
  learning. pp. 448--456. pmlr (2015)

\bibitem{kingma2014adam}
Kingma, D.P., Ba, J.: Adam: A method for stochastic optimization. arXiv
  preprint arXiv:1412.6980  (2014)

\bibitem{krull2019noise2void}
Krull, A., Buchholz, T.O., Jug, F.: Noise2void-learning denoising from single
  noisy images. In: Proceedings of the IEEE/CVF conference on computer vision
  and pattern recognition. pp. 2129--2137 (2019)

\bibitem{laine2019high}
Laine, S., Karras, T., Lehtinen, J., Aila, T.: High-quality self-supervised
  deep image denoising. Advances in Neural Information Processing Systems
  \textbf{32} (2019)

\bibitem{lee2021restore}
Lee, S., Cho, D., Kim, J., Kim, T.H.: Restore from restored: Video restoration
  with pseudo clean video. In: Proceedings of the IEEE/CVF Conference on
  Computer Vision and Pattern Recognition. pp. 3537--3546 (2021)

\bibitem{lehtinen2018noise2noise}
Lehtinen, J., Munkberg, J., Hasselgren, J., Laine, S., Karras, T., Aittala, M.,
  Aila, T.: Noise2noise: Learning image restoration without clean data. arXiv
  preprint arXiv:1803.04189  (2018)

\bibitem{li2023simple}
Li, D., Shi, X., Zhang, Y., Cheung, K.C., See, S., Wang, X., Qin, H., Li, H.: A
  simple baseline for video restoration with grouped spatial-temporal shift.
  In: Proceedings of the IEEE/CVF Conference on Computer Vision and Pattern
  Recognition. pp. 9822--9832 (2023)

\bibitem{li2022unidirectional}
Li, J., Wu, X., Niu, Z., Zuo, W.: Unidirectional video denoising by mimicking
  backward recurrent modules with look-ahead forward ones. In: European
  Conference on Computer Vision. pp. 592--609. Springer (2022)

\bibitem{liang2022vrt}
Liang, J., Cao, J., Fan, Y., Zhang, K., Ranjan, R., Li, Y., Timofte, R.,
  Van~Gool, L.: Vrt: A video restoration transformer. arXiv preprint
  arXiv:2201.12288  (2022)

\bibitem{liang2021swinir}
Liang, J., Cao, J., Sun, G., Zhang, K., Van~Gool, L., Timofte, R.: Swinir:
  Image restoration using swin transformer. In: Proceedings of the IEEE/CVF
  international conference on computer vision. pp. 1833--1844 (2021)

\bibitem{liang2022recurrent}
Liang, J., Fan, Y., Xiang, X., Ranjan, R., Ilg, E., Green, S., Cao, J., Zhang,
  K., Timofte, R., Gool, L.V.: Recurrent video restoration transformer with
  guided deformable attention. Advances in Neural Information Processing
  Systems  \textbf{35},  378--393 (2022)

\bibitem{lim2017enhanced}
Lim, B., Son, S., Kim, H., Nah, S., Mu~Lee, K.: Enhanced deep residual networks
  for single image super-resolution. In: Proceedings of the IEEE conference on
  computer vision and pattern recognition workshops. pp. 136--144 (2017)

\bibitem{liu2018non}
Liu, D., Wen, B., Fan, Y., Loy, C.C., Huang, T.S.: Non-local recurrent network
  for image restoration. Advances in neural information processing systems
  \textbf{31} (2018)

\bibitem{loshchilov2016sgdr}
Loshchilov, I., Hutter, F.: Sgdr: Stochastic gradient descent with warm
  restarts. arXiv preprint arXiv:1608.03983  (2016)

\bibitem{ma2016waterloo}
Ma, K., Duanmu, Z., Wu, Q., Wang, Z., Yong, H., Li, H., Zhang, L.: Waterloo
  exploration database: New challenges for image quality assessment models.
  IEEE Transactions on Image Processing  \textbf{26}(2),  1004--1016 (2016)

\bibitem{maggioni2012video}
Maggioni, M., Boracchi, G., Foi, A., Egiazarian, K.: Video denoising,
  deblocking, and enhancement through separable 4-d nonlocal spatiotemporal
  transforms. IEEE Transactions on image processing  \textbf{21}(9),
  3952--3966 (2012)

\bibitem{maggioni2021efficient}
Maggioni, M., Huang, Y., Li, C., Xiao, S., Fu, Z., Song, F.: Efficient
  multi-stage video denoising with recurrent spatio-temporal fusion. In:
  Proceedings of the IEEE/CVF Conference on Computer Vision and Pattern
  Recognition. pp. 3466--3475 (2021)

\bibitem{martin2001database}
Martin, D., Fowlkes, C., Tal, D., Malik, J.: A database of human segmented
  natural images and its application to evaluating segmentation algorithms and
  measuring ecological statistics. In: Proceedings Eighth IEEE International
  Conference on Computer Vision. ICCV 2001. vol.~2, pp. 416--423. IEEE (2001)

\bibitem{mildenhall2018burst}
Mildenhall, B., Barron, J.T., Chen, J., Sharlet, D., Ng, R., Carroll, R.: Burst
  denoising with kernel prediction networks. In: Proceedings of the IEEE
  conference on computer vision and pattern recognition. pp. 2502--2510 (2018)

\bibitem{nair2010rectified}
Nair, V., Hinton, G.E.: Rectified linear units improve restricted boltzmann
  machines. In: Proceedings of the 27th international conference on machine
  learning (ICML-10). pp. 807--814 (2010)

\bibitem{nam2016holistic}
Nam, S., Hwang, Y., Matsushita, Y., Kim, S.J.: A holistic approach to
  cross-channel image noise modeling and its application to image denoising.
  In: Proceedings of the IEEE conference on computer vision and pattern
  recognition. pp. 1683--1691 (2016)

\bibitem{pang2021recorrupted}
Pang, T., Zheng, H., Quan, Y., Ji, H.: Recorrupted-to-recorrupted: Unsupervised
  deep learning for image denoising. In: Proceedings of the IEEE/CVF conference
  on computer vision and pattern recognition. pp. 2043--2052 (2021)

\bibitem{plotz2017benchmarking}
Plotz, T., Roth, S.: Benchmarking denoising algorithms with real photographs.
  In: Proceedings of the IEEE conference on computer vision and pattern
  recognition. pp. 1586--1595 (2017)

\bibitem{pont20172017}
Pont-Tuset, J., Perazzi, F., Caelles, S., Arbel{\'a}ez, P., Sorkine-Hornung,
  A., Van~Gool, L.: The 2017 davis challenge on video object segmentation.
  arXiv preprint arXiv:1704.00675  (2017)

\bibitem{sheth2021unsupervised}
Sheth, D.Y., Mohan, S., Vincent, J.L., Manzorro, R., Crozier, P.A., Khapra,
  M.M., Simoncelli, E.P., Fernandez-Granda, C.: Unsupervised deep video
  denoising. In: Proceedings of the IEEE/CVF International Conference on
  Computer Vision. pp. 1759--1768 (2021)

\bibitem{song2022tempformer}
Song, M., Zhang, Y., Ayd{\i}n, T.O.: Tempformer: Temporally consistent
  transformer for video denoising. In: European Conference on Computer Vision.
  pp. 481--496. Springer (2022)

\bibitem{sun2018pwc}
Sun, D., Yang, X., Liu, M.Y., Kautz, J.: Pwc-net: Cnns for optical flow using
  pyramid, warping, and cost volume. In: Proceedings of the IEEE conference on
  computer vision and pattern recognition. pp. 8934--8943 (2018)

\bibitem{tassano2019dvdnet}
Tassano, M., Delon, J., Veit, T.: Dvdnet: A fast network for deep video
  denoising. In: 2019 IEEE International Conference on Image Processing (ICIP).
  pp. 1805--1809. IEEE (2019)

\bibitem{tassano2020fastdvdnet}
Tassano, M., Delon, J., Veit, T.: Fastdvdnet: Towards real-time deep video
  denoising without flow estimation. In: Proceedings of the IEEE/CVF conference
  on computer vision and pattern recognition. pp. 1354--1363 (2020)

\bibitem{vaksman2021patch}
Vaksman, G., Elad, M., Milanfar, P.: Patch craft: Video denoising by deep
  modeling and patch matching. In: Proceedings of the IEEE/CVF International
  Conference on Computer Vision. pp. 2157--2166 (2021)

\bibitem{wang2024progressive}
Wang, C., Guo, L., Wang, Y., Cheng, H., Yu, Y., Wen, B.: Progressive
  divide-and-conquer via subsampling decomposition for accelerated mri. In:
  Proceedings of the IEEE/CVF Conference on Computer Vision and Pattern
  Recognition. pp. 25128--25137 (2024)

\bibitem{wang2019edvr}
Wang, X., Chan, K.C., Yu, K., Dong, C., Change~Loy, C.: Edvr: Video restoration
  with enhanced deformable convolutional networks. In: Proceedings of the
  IEEE/CVF conference on computer vision and pattern recognition workshops.
  pp.~0--0 (2019)

\bibitem{wang2022uformer}
Wang, Z., Cun, X., Bao, J., Zhou, W., Liu, J., Li, H.: Uformer: A general
  u-shaped transformer for image restoration. In: Proceedings of the IEEE/CVF
  conference on computer vision and pattern recognition. pp. 17683--17693
  (2022)

\bibitem{wang2023recurrent}
Wang, Z., Zhang, Y., Zhang, D., Fu, Y.: Recurrent self-supervised video
  denoising with denser receptive field. In: Proceedings of the 31st ACM
  International Conference on Multimedia. pp. 7363--7372 (2023)

\bibitem{wei2020physics}
Wei, K., Fu, Y., Yang, J., Huang, H.: A physics-based noise formation model for
  extreme low-light raw denoising. In: Proceedings of the IEEE/CVF Conference
  on Computer Vision and Pattern Recognition. pp. 2758--2767 (2020)

\bibitem{wen2018vidosat}
Wen, B., Ravishankar, S., Bresler, Y.: Vidosat: High-dimensional sparsifying
  transform learning for online video denoising. IEEE Transactions on Image
  processing  \textbf{28}(4),  1691--1704 (2018)

\bibitem{wu2020unpaired}
Wu, X., Liu, M., Cao, Y., Ren, D., Zuo, W.: Unpaired learning of deep image
  denoising. In: European conference on computer vision. pp. 352--368. Springer
  (2020)

\bibitem{xu2018real}
Xu, J., Li, H., Liang, Z., Zhang, D., Zhang, L.: Real-world noisy image
  denoising: A new benchmark. arXiv preprint arXiv:1804.02603  (2018)

\bibitem{xue2019video}
Xue, T., Chen, B., Wu, J., Wei, D., Freeman, W.T.: Video enhancement with
  task-oriented flow. International Journal of Computer Vision  \textbf{127},
  1106--1125 (2019)

\bibitem{yue2020supervised}
Yue, H., Cao, C., Liao, L., Chu, R., Yang, J.: Supervised raw video denoising
  with a benchmark dataset on dynamic scenes. In: Proceedings of the IEEE/CVF
  conference on computer vision and pattern recognition. pp. 2301--2310 (2020)

\bibitem{zamir2022restormer}
Zamir, S.W., Arora, A., Khan, S., Hayat, M., Khan, F.S., Yang, M.H.: Restormer:
  Efficient transformer for high-resolution image restoration. In: Proceedings
  of the IEEE/CVF conference on computer vision and pattern recognition. pp.
  5728--5739 (2022)

\bibitem{zhang2021plug}
Zhang, K., Li, Y., Zuo, W., Zhang, L., Van~Gool, L., Timofte, R.: Plug-and-play
  image restoration with deep denoiser prior. IEEE Transactions on Pattern
  Analysis and Machine Intelligence  \textbf{44}(10),  6360--6376 (2021)

\bibitem{zhang2017beyond}
Zhang, K., Zuo, W., Chen, Y., Meng, D., Zhang, L.: Beyond a gaussian denoiser:
  Residual learning of deep cnn for image denoising. IEEE transactions on image
  processing  \textbf{26}(7),  3142--3155 (2017)

\bibitem{zhang2022idr}
Zhang, Y., Li, D., Law, K.L., Wang, X., Qin, H., Li, H.: Idr: Self-supervised
  image denoising via iterative data refinement. In: Proceedings of the
  IEEE/CVF Conference on Computer Vision and Pattern Recognition. pp.
  2098--2107 (2022)

\bibitem{zhang2021rethinking}
Zhang, Y., Qin, H., Wang, X., Li, H.: Rethinking noise synthesis and modeling
  in raw denoising. In: Proceedings of the IEEE/CVF International Conference on
  Computer Vision. pp. 4593--4601 (2021)

\bibitem{zheng2023unsupervised}
Zheng, H., Pang, T., Ji, H.: Unsupervised deep video denoising with untrained
  network. In: Proceedings of the AAAI Conference on Artificial Intelligence.
  vol.~37, pp. 3651--3659 (2023)

\end{thebibliography}

\clearpage
\renewcommand\thesection{\Alph{section}}
\setcounter{section}{0}

\section{Architecture of Image Denoising Baseline}
We use NAFNet~\cite{chen2022simple}, a 4-level encoder-decoder, as our image denoising baseline. We use the residual blocks introduced in~\cite{chen2022simple}. The number of residual blocks in the encoder from level-1 to level-4 is [2, 2, 4, 6], with the channel number set to [64, 128, 256, 512] correspondingly. In the decoder, the number of residual blocks from level-3 to level-1 is [2, 2, 2]. 

\section{Selection of Noise Models}
During fine-tuning, the noise sampled from the noise model is utilized to corrupt the pseudo clean video. For synthetic Gaussian video denoising, we use additive white Gaussian noise (AWGN):
\begin{equation}
    n_i \sim \mathcal{N}(0,\sigma^2),
\end{equation}
where $n_i$ denotes the noise value at pixel $i$, and $\sigma$ is the standard deviation of AWGN. 
For a noisy video with noise level $\sigma$, we first denoise the video with our pre-trained blind Gaussian denoiser. Subsequently, noise sampled from the AWGN at the same noise level $\sigma$ is added to the denoised video to construct pseudo noisy-clean video pairs. 
For real raw video denoising, we use Poisson-Gaussian noise:
\begin{equation}
    n_i \sim \alpha\mathcal{P}(x_i) - x_i + \mathcal{N}(0,\delta^2),
\end{equation}
where $\mathcal{P}$ denotes the Poisson noise, $x_i$ is the image intensity at pixel $i$, and $\alpha, \delta$ are parameters related to ISO. These parameters can be calibrated following previous works~\cite{wei2020physics, zhang2021rethinking}. In particular, the CRVD dataset~\cite{yue2020supervised} provides the camera parameters for various ISO levels, which can be directly utilized to synthesize the noise to the pseudo clean video.

\section{Visual Examples}
\begin{figure}[!t]
    \centering
    \includegraphics[width=\linewidth]{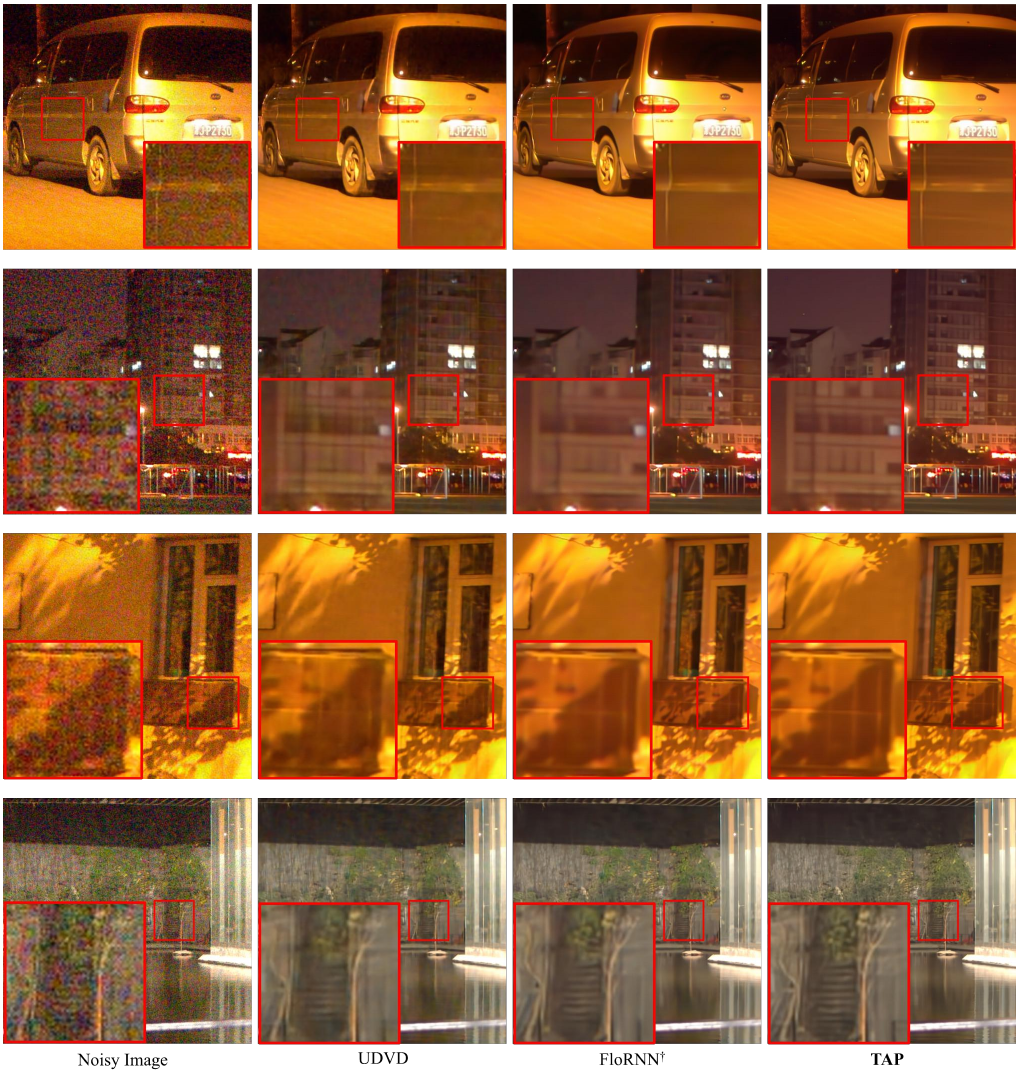}
    \caption{Visual examples of real raw video denoising on CRVD outdoor set~\cite{yue2020supervised}. The noisy image, the restored images of UDVD~\cite{sheth2021unsupervised}, FloRNN~\cite{li2022unidirectional}, and TAP, respectively. \dag~denotes the supervised method. We render raw images to sRGB images with the pre-trained ISP provided in~\cite{yue2020supervised} (ISO$=$25600).}
    \label{fig_raw}
\end{figure}

\begin{figure}[!t]
    \centering
    \includegraphics[width=\linewidth]{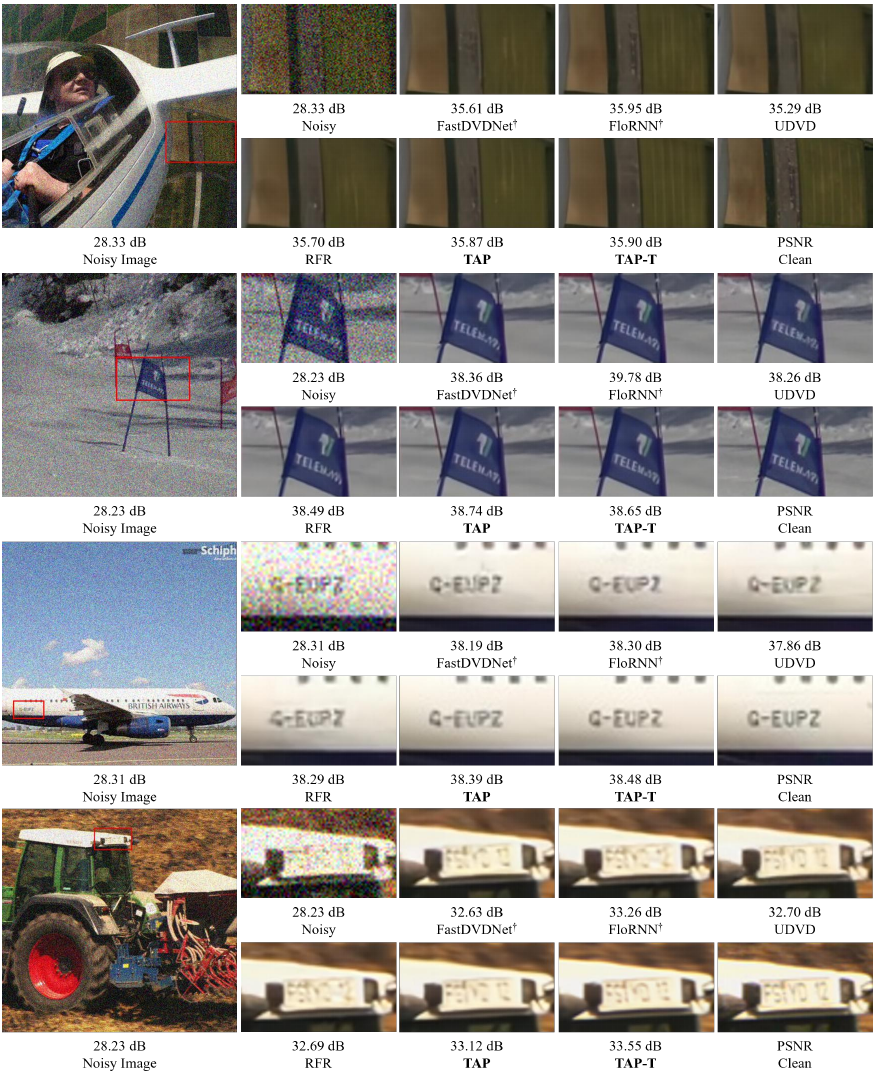}
    \caption{Visual examples of synthetic Gaussian video denoising on DAVIS~\cite{pont20172017} and Set8~\cite{tassano2020fastdvdnet} datasets, including the noisy inputs (with noise level $\sigma=30$), the restored images using FastDVDNet~\cite{tassano2020fastdvdnet}, FloRNN~\cite{li2022unidirectional}, UDVD~\cite{sheth2021unsupervised}, RFR~\cite{lee2021restore}, TAP, and TAP-T, as well as the clean ground-truth, respectively. \dag~indicates the supervised method.}
    \label{fig_rgb}
\end{figure}

\begin{itemize}
    \renewcommand\labelitemi{$\bullet$}
    \item 
    Real raw video denoising:~\cref{fig_raw}.

    \item
    Synthetic Gaussian video denoising:~\cref{fig_rgb}
    
\end{itemize}

\end{document}